%% file: main.tex
\documentclass[10pt,twocolumn,letterpaper]{article}

\usepackage[pagenumbers]{cvpr} 

\input{preamble}

\definecolor{cvprblue}{rgb}{0.21,0.49,0.74}
\usepackage[pagebackref,breaklinks,colorlinks,allcolors=cvprblue,hypertexnames=false]{hyperref}

\def\confName{CVPR}
\def\confYear{2027}

\title{$R^2$: Recognizing and Reducing Redundant Visual Tool Use}

\author{%
Yilin Zhao\thanks{Equal contribution.}\quad
Jinyu Wang\footnotemark[1]\quad
Wenming Yang\thanks{Corresponding author.}\\
Tsinghua University\\
{\tt\small z-yl24@mails.tsinghua.edu.cn, jywang24@mails.tsinghua.edu.cn}\\
{\tt\small yang.wenming@sz.tsinghua.edu.cn}
}

\begin{document}
\maketitle
\suppressfloats[t]
\input{sec/0_abstract}
\input{sec/teaser_placeholder}    
\input{sec/1_intro}
\input{sec/2_related}
\input{sec/method_placeholder}
\input{sec/3_method}
\input{sec/4_experiments}

\input{sec/5_conclusion}
\flushbottom
\input{sec/llm_usage}
{
    \small
    \bibliographystyle{ieeenat_fullname_citationorder}
    \bibliography{main}
}


\clearpage
\appendix
\setcounter{page}{1}
\setcounter{dbltopnumber}{1}
\renewcommand{\dbltopfraction}{0.9}
\makeatletter
\setlength{\@dblfptop}{0pt}
\makeatother
\input{sec/X_vfd_validation}
\input{sec/X_cat_details}
\input{sec/X_experimental_setup}
\input{sec/X_ablation_details}
\input{sec/X_vfd_other_tools}

\end{document}

%% file: preamble.tex
\usepackage{booktabs}
\usepackage{xcolor}
\usepackage{colortbl}

\usepackage{microtype}

%% file: sec/0_abstract.tex
\begin{abstract}
Recent studies have shown that visual tools can improve multimodal reasoning. However, the effects of tools vary substantially across samples, and benchmark accuracy alone does not reveal whether tool interactions are beneficial or redundant. We present $R^2$, a framework that recognizes and reduces redundant visual tool use in vision-language models. We first introduce a sample-level Category Analysis of Tool Utility (CAT), which categorizes samples according to whether a model can benefit from tools. This analysis reveals that benchmark scores may obscure a model’s tool-use ability. Across seven benchmarks, samples benefiting from tools account for at most 12.44\% of each benchmark, while the majority gain little from tool interaction or are even harmed by it. To assess individual calls, we propose the Visual Feedback Dependency (VFD) metric, which compares the likelihood of the same subsequent reasoning under actual and alternative visual feedback without an external evaluator or trajectory regeneration. Redundancy-Aware GRPO (RA-GRPO) combines this dependency signal with a category-conditioned base call cost to assign relative efficiency advantages among correct trajectories, supplementing outcome-based supervision. Across the seven benchmarks, RA-GRPO reduces the mean number of tool calls by 12.39\% relative to Vanilla GRPO at comparable benchmark accuracy. Evaluated by tool-utility category, RA-GRPO improves category macro-average accuracy over Vanilla GRPO by 3.35 percentage points.
\end{abstract}

%% file: sec/teaser_placeholder.tex
\begin{figure}[t]
\centering
\includegraphics[width=\columnwidth]{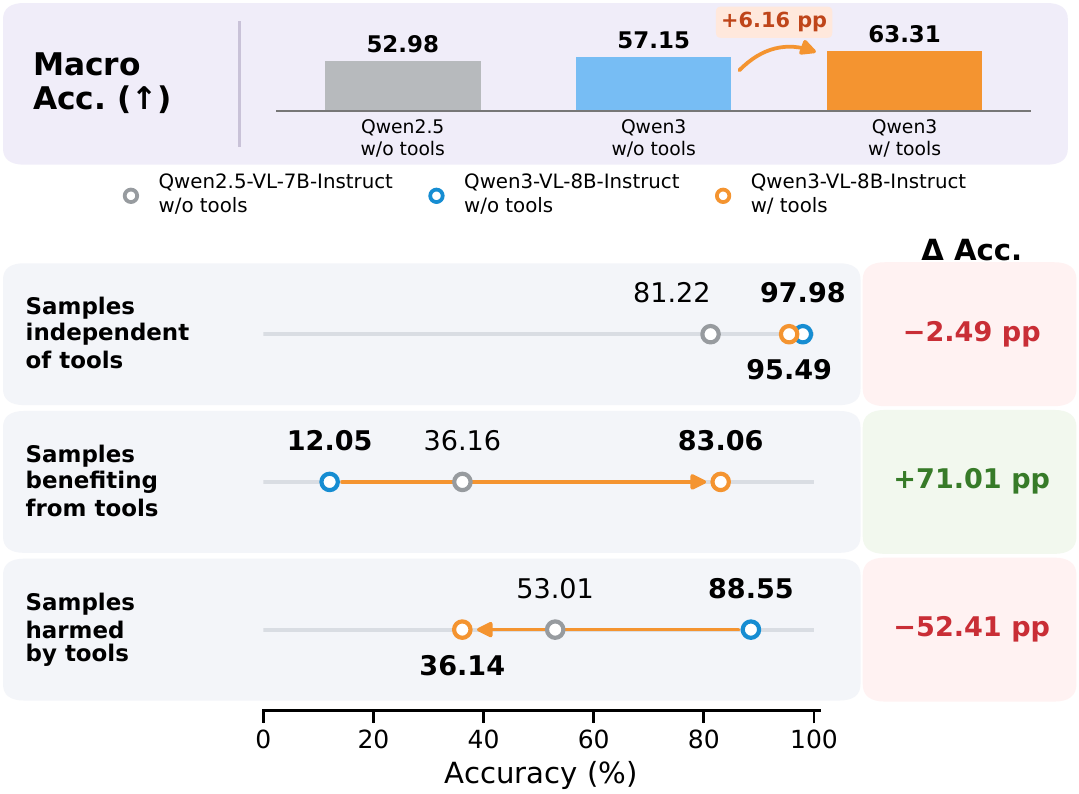}
\caption{\textbf{Benchmark macro-average accuracy obscures large performance differences between Qwen3-VL-8B-Instruct with and without tool access on samples that benefit from or are harmed by tools.} Enabling tools improves macro-average accuracy by 6.16 percentage points, yet Category Analysis of Tool Utility reveals sharply contrasting results on these samples.}
\label{fig:teaser_placeholder}
\vspace{-6pt}
\end{figure}

%% file: sec/1_intro.tex
\section{Introduction}
\label{sec:intro}

Visual tools enable vision-language models (VLMs) to gather additional evidence during reasoning. In the thinking-with-images paradigm, models interleave textual reasoning with operations such as cropping, zooming, and rotation to inspect fine-grained visual details~\cite{openai_thinking_images}. Building on advances in reinforcement learning with verifiable rewards~\cite{deepseekr1,dapo}, multimodal reasoning methods adopt GRPO~\cite{deepseekmath} to develop these capabilities~\cite{deepeyes,chain_of_focus,minio3,treevgr}. Recent work further extends visual interaction to executable code and external APIs~\cite{deepeyesv2,thyme}. As tool use becomes a central part of visual reasoning, a natural question is whether each tool interaction actually contributes to the answer.

However, higher answer accuracy does not necessarily indicate a better tool-use strategy. The benefits of visual tools vary across samples. Some can be solved without tools, some benefit from additional observations, and others exhibit lower accuracy when tools are available. Figure~\ref{fig:teaser_placeholder} illustrates this discrepancy. Enabling tools improves the benchmark macro-average accuracy of Qwen3-VL-8B-Instruct by 6.16 percentage points, yet this aggregate gain masks sharply contrasting effects on samples that benefit from tools and those harmed by them. Moreover, even when a sample benefits from tools, individual calls may provide repeated or low-value observations. Effective tool-use supervision therefore requires distinguishing sample-level tool utility from the value of individual calls. Outcome rewards alone do not make this distinction, since equally correct trajectories receive the same reward regardless of differences in their tool-use efficiency.

We introduce $R^2$, a framework connecting sample-level tool-utility analysis, turn-level feedback diagnosis, and redundancy-aware policy optimization. We first propose Category Analysis of Tool Utility (CAT), which compares teacher-model rollouts with and without tool access to distinguish five categories, namely Tool-Independent, Tool-Beneficial, Challenging, Tool-Harmful, and Ambiguous. CAT characterizes tool-utility distributions across benchmarks and guides training-data selection. Our analysis of seven benchmarks finds that Tool-Beneficial samples account for at most 12.44\% of each benchmark, while more than 60\% of samples in HRBench and V$^\star$Bench are Tool-Independent, as shown in Figure~\ref{fig:cat_distribution}. These findings explain why aggregate scores alone provide an incomplete view of tool-use performance. 

To assess individual calls, we propose Visual Feedback Dependency (VFD), a turn-level metric that compares the conditional log-likelihood of the same subsequent reasoning under actual and alternative visual feedback. VFD measures how strongly the generated reasoning depends on a particular observation, providing a signal of potential redundancy without an external evaluator or trajectory regeneration. We then propose Redundancy-Aware GRPO (RA-GRPO), which combines VFD-based redundancy estimation with a base call cost applied to Tool-Independent samples. RA-GRPO assigns relative efficiency advantages among correct trajectories, supplying an efficiency signal where outcome advantages are identical without rewarding low-cost failures.

Across seven benchmarks, our model improves macro-average accuracy from 63.31\% to 68.13\% over the tool-enabled Qwen3-VL-8B-Instruct backbone while reducing mean tool calls by 15.38\%. Category-wise evaluation reveals where these changes occur. Relative to Vanilla GRPO, RA-GRPO improves Tool-Beneficial accuracy while reducing calls by 12.28\%, showing that samples benefiting from tools can still be solved with fewer calls. Its largest accuracy gain occurs on Tool-Harmful samples, with an improvement of 13.86 percentage points and a 25.19\% reduction in calls. Together, these results demonstrate the value of examining performance by tool utility rather than relying on aggregate benchmark accuracy alone. Our contributions are summarized below.

\begin{itemize}
  \item \textbf{Sample-level tool-utility analysis.} We introduce CAT to characterize tool-utility distributions and reveal category-specific strengths and weaknesses obscured by aggregate benchmark scores, informing benchmark analysis and training-data selection.
  \item \textbf{Turn-level feedback diagnosis.} We propose VFD to measure the dependence of subsequent reasoning on individual visual observations without an external evaluator or trajectory regeneration. Offline validation supports its use as a signal for identifying potentially redundant calls.
  \item \textbf{Redundancy-aware policy optimization.} We develop RA-GRPO to combine VFD-based redundancy costs with a category-conditioned base cost and assign relative efficiency advantages among correct trajectories. Relative to Vanilla GRPO, RA-GRPO reduces mean tool calls by 12.39\% at comparable benchmark accuracy and improves category macro-average accuracy by 3.35 percentage points.
\end{itemize}

%% file: sec/2_related.tex
\section{Related Work}
\label{sec:related}

\subsection{Thinking with Images}
Recent work equips VLMs with visual operations, such as cropping and zooming tools. DeepEyes~\cite{deepeyes} learns this capability directly through reinforcement learning without cold-start SFT, while Pixel-Reasoner~\cite{pixelreasoner} combines instruction tuning with curiosity-driven rewards to encourage the exploration of visual operations. Chain-of-Focus~\cite{chain_of_focus} learns to adapt its visual search and zooming, and Mini-o3~\cite{minio3} increases the diversity and depth of multi-turn visual search. Thyme~\cite{thyme} extends visual interaction to executable image processing and computation, and PyVision-RL~\cite{pyvisionrl} mitigates interaction collapse through rollout selection and accumulative tool rewards. Most of these methods have not yet focused on whether the samples themselves depend on tools. Our work addresses this gap by studying how to reduce repeated or low-value observations while retaining the model capability gains brought by tool interaction.

\subsection{Tool Utility and Credit Assignment}
Another line of work questions whether accuracy gains actually come from visual tools. Guo et al. and Ma et al. \cite{guo_tool_benefit,ma_tool_learning} show that the benefit of tool access itself is limited, and most of the improvement comes from intrinsic learning and reduced tool-induced harm. TAPO in CodeV rewards evidence in intermediate tool outputs \cite{codev}. TACO estimates call contribution through answer probes with and without a tool, then routes advantages to relevant segments \cite{taco}. CounterCredit uses gold-answer scores to evaluate the value of looking and the value of returned evidence, including comparisons with random replacement crops \cite{countercredit}. These findings motivate us to separate sample-level tool utility from the diagnosis of turn-level invocation redundancy. We use CAT to reveal the tool-utility sample distribution of the benchmarks and filter training examples; VFD then measures how subsequent reasoning depends on a specific visual observation, and RA-GRPO converts this dependence signal into estimated redundancy costs and an efficiency advantage within correct trajectories.

%% file: sec/method_placeholder.tex
\begin{figure*}[t]
\centering
\includegraphics[width=0.92\textwidth]{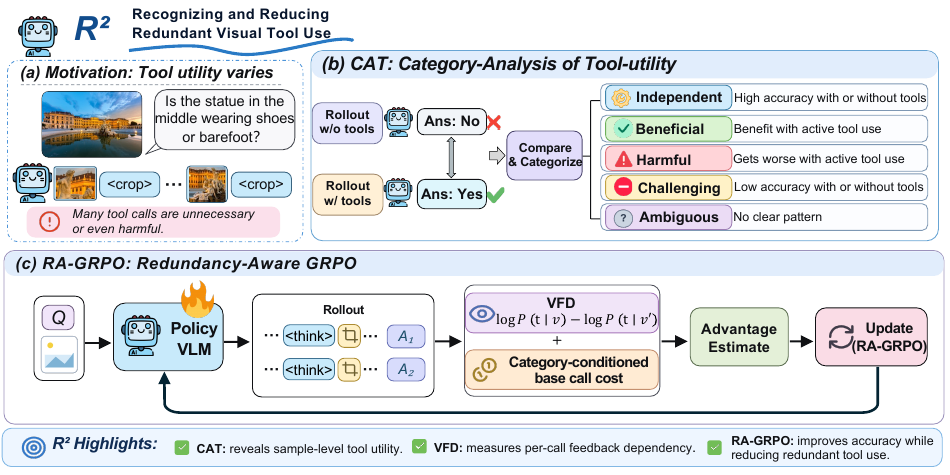}
\caption{\textbf{Overview of $R^2$ for recognizing and reducing redundant visual tool use.} CAT compares rollouts with and without tool access to characterize sample-level tool utility and guide training-data selection. VFD measures the dependence of the same subsequent reasoning on actual versus alternative visual feedback. RA-GRPO combines VFD-based redundancy costs with a base call cost for Tool-Independent samples to assign relative efficiency advantages among correct trajectories, supplementing outcome supervision.}
\label{fig:method_placeholder}
\end{figure*}

%% file: sec/3_method.tex
\section{Method}
\label{sec:method}
We introduce $R^2$, a framework designed to recognize and reduce redundant visual tool use while improving model performance. Category Analysis of Tool Utility (CAT) characterizes how each sample responds to tool access and guides training-data selection, as described in Sec.~\ref{sec:cat}. Redundancy-Aware GRPO (RA-GRPO) then uses Visual Feedback Dependency (VFD) to estimate redundant-call costs and combines these estimates with a category-conditioned base cost to favor more efficient correct trajectories, as described in Sec.~\ref{sec:ra_grpo}.

\subsection{Category Analysis of Tool Utility}
\label{sec:cat}

Task difficulty alone does not indicate whether a sample benefits from visual tools.  We introduce Category Analysis of Tool Utility (CAT) to characterize these differences and guide benchmark analysis and training-data selection.

For each sample, we compare multiple teacher-model rollouts under two settings, namely tool-enabled reasoning, which permits multi-turn tool interaction, and tool-disabled reasoning, which uses only the original image as visual input. CAT considers both answer correctness and whether tools are actually invoked to characterize sample-level tool utility and its distribution across datasets and benchmarks.

We distinguish five categories. \emph{Tool-Independent} samples are reliably solved both with and without tool access, indicating that direct reasoning is sufficient. \emph{Tool-Beneficial} samples exhibit substantially higher success under tool-enabled reasoning with active tool use. \emph{Challenging} samples have low success in both settings, indicating that tool access alone does not resolve their difficulty. \emph{Tool-Harmful} samples exhibit lower success under tool-enabled reasoning with active tool use. \emph{Ambiguous} samples show intermediate or inconclusive patterns that do not meet the criteria of the preceding categories. Detailed classification rules are provided in Appendix~\ref{sec:cat_classification_rules}. We repeat CAT with Mini-o3~\cite{minio3} as the tool-enabled teacher. The exact proportions shift, but the dominant category on each benchmark is preserved. Tool-Beneficial remains a minority throughout, as reported in Appendix~\ref{sec:cat_teacher_comparison}.

Sample-level utility does not determine the value of every call. Even a Tool-Beneficial sample may elicit repeated or low-value observations. 

\subsection{Redundancy-Aware GRPO}
\label{sec:ra_grpo}
Redundancy-Aware GRPO (RA-GRPO) uses Visual Feedback Dependency (VFD) to assess how subsequent reasoning depends on individual tool observations and combines this signal with CAT to estimate trajectory costs. It then assigns relative efficiency advantages among correct trajectories, supplementing correctness supervision to reduce low-value visual observations.

Given an image-question input $x$, we sample a group of $G$ trajectories $\{o_i\}_{i=1}^{G}$ from the old policy $\pi_{\theta_{\mathrm{old}}}$ where $i$ indexes trajectories. Each trajectory contains generated text, tool calls, and the observations returned by those calls. A trajectory receives a reward of $r_i=1$ when its answer is correct and its format is valid, and $0$ otherwise.

Following GRPO \cite{deepseekmath}, we compute the outcome advantage as
\begin{equation}
A_i^{\mathrm{out}}=\frac{r_i-\bar r}{\sigma_r+\epsilon},
\qquad
\bar r=\frac{1}{G}\sum_{j=1}^{G}r_j,
\label{eq:outcome_advantage}
\end{equation}
where $\sigma_r$ is the sample standard deviation of rewards within the group, and $\epsilon$ is a small positive constant. Within a group, trajectories receiving a reward of \(1\) have identical outcome advantages, which do not distinguish differences in their tool-use efficiency. 

\subsubsection{Visual Feedback Dependency}
\label{sec:vfd}
Supervising whether or how often tools are invoked does not, by itself, capture the information value of the returned observations. VFD assesses feedback dependency by comparing the conditional likelihood of the same subsequent reasoning text under actual and alternative visual feedback.

For the $t$-th tool call in trajectory $o_i$, let $v_{i,t}$ denote the actual visual feedback and $h_{i,t}$ the complete context preceding that feedback. Let $u_{i,t}=(u_1,\ldots,u_L)$ denote the $L$-token reasoning window following the feedback and ending before the next tool call or the final answer. For a reasoning window $u$, we define its length-normalized conditional log-likelihood as
\begin{equation}
\ell(u\mid h,v)=\frac{1}{L}\sum_{k=1}^{L}
\log\pi_{\theta_{\mathrm{old}}}(u_k\mid h,v,u_{<k}),
\label{eq:conditional_likelihood}
\end{equation}
where $u_{<k}$ denotes the preceding tokens in the window.
We replace only the visual feedback from the current call, keeping the original image, the remaining history, and the target text unchanged. Given $J$ alternative observations $\{\tilde v_{i,t}^{(j)}\}_{j=1}^{J}$, we define the VFD score as
\begin{equation}
\begin{aligned}
d_{i,t}={}&\ell(u_{i,t}\mid h_{i,t},v_{i,t})\\
&-\frac{1}{J}\sum_{j=1}^{J}\ell(u_{i,t}\mid h_{i,t},\tilde v_{i,t}^{(j)}).
\end{aligned}
\label{eq:vfd}
\end{equation}
The score measures the additional log-likelihood support that actual feedback provides for the generated reasoning relative to alternative feedback. Higher scores indicate stronger dependence on the current observation; lower scores serve as a signal of potential redundancy. For cropping tools, we construct alternative observations by randomly sampling regions from the source image of the current call. Each region has the same size as the actual crop and does not overlap with it. All comparisons score the same already-generated reasoning text and require no regeneration of trajectories under alternative visual feedback. We evaluate VFD offline against call-level necessity annotations and report the protocol and results in Appendix~\ref{sec:vfd_offline_validation}.

To account for differences in VFD distributions across data sources, we calibrate a separate threshold for each source using reasoning windows collected during training warmup as follows.
\begin{equation}
\tau_s=\operatorname{Quantile}_{q}\!\left(
\left\{d_{i,t}:\begin{array}{l}s(x)=s,\\(i,t)\in\mathcal W_{\mathrm{warm}}^{\mathrm{reason}}\end{array}\right\}
\right),
\label{eq:vfd_threshold}
\end{equation}
where $s(x)$ denotes the data source of the sample, $\mathcal W_{\mathrm{warm}}^{\mathrm{reason}}$ denotes the collection of warmup reasoning windows, and $q$ is the quantile level. After warmup, we freeze the thresholds and mark calls with $d_{i,t}<\tau_{s(x)}$ as low-dependency calls.

\subsubsection{Redundancy-Aware Advantages and Policy Optimization}
\label{sec:ra_advantage}
\input{sec/table_main_results}

\paragraph{Trajectory cost.}
To limit online computational overhead, we uniformly sample one eligible call $t_i$ per trajectory. A call is eligible if it successfully returns a crop and meets the requirements for VFD scoring. Let $\mathcal E_i$ denote the set of eligible calls and $M_i=|\mathcal E_i|$. We estimate the number of low-dependency calls among the eligible calls as
\begin{equation}
\begin{aligned}
\widehat R_i&=M_i z_i,\qquad t_i\sim\operatorname{Uniform}(\mathcal E_i),\\
z_i&=\mathbf 1\!\left[d_{i,t_i}<\tau_{s(x)}\right].
\end{aligned}
\label{eq:redundant_calls}
\end{equation}
If $M_i=0$, we skip sampling and set $\widehat R_i=0$. Further details are provided in Appendix~\ref{sec:vfd_implementation}.

The VFD-based cost estimate alone may assign zero cost both to direct-answer trajectories and to tool-using trajectories whose sampled calls are not identified as low-dependency. To further encourage the model to answer directly on Tool-Independent samples, we define the category-conditioned trajectory cost as
\begin{equation}
C_i=\widehat R_i+K\,\mathbf 1[c(x)=\text{Tool-Independent}]\,N_i,
\label{eq:trajectory_cost}
\end{equation}
where $c(x)$ is the CAT category assigned before training, $N_i$ is the total number of tool calls in trajectory $o_i$, and $K\geq0$ is the base cost per call. Direct-answer trajectories have zero cost. On Tool-Independent samples, any trajectory with at least one tool call has a positive cost whenever $K>0$. For the remaining categories, trajectory costs consist only of the VFD-based redundancy estimate.

\paragraph{Efficiency advantage.}
To avoid rewarding low-cost failures through the efficiency term, we compare costs only among trajectories receiving $r_i=1$ within the same group. Their index set is $\mathcal S^+=\{i\in\{1,\ldots,G\}:r_i=1\}$. When $\mathcal S^+\neq\varnothing$, let $\bar C^+=|\mathcal S^+|^{-1}\sum_{i\in\mathcal S^+}C_i$ denote their mean cost. We define
\begin{equation}
A_i^{\mathrm{eff}}=
\begin{cases}
\dfrac{\bar C^+-C_i}{B},&i\in\mathcal S^+,\\[5pt]
0,&i\notin\mathcal S^+.
\end{cases}
\label{eq:efficiency_advantage}
\end{equation}
We normalize cost differences by the fixed maximum tool-call budget \(B\), rather than the within-group cost standard deviation. If $\mathcal S^+=\varnothing$, or if all correct trajectories have the same cost, the efficiency advantages in the group are zero.

We combine the outcome and efficiency advantages as
\begin{equation}
A_i=A_i^{\mathrm{out}}+\beta A_i^{\mathrm{eff}},
\label{eq:combined_advantage}
\end{equation}
where $\beta\geq0$ controls the contribution of the efficiency advantage. Within the correct set, efficiency advantages sum to zero. Trajectories with below-average costs receive positive efficiency advantages, whereas those with above-average costs receive negative ones. 

For each generated token $o_{i,m}$ in trajectory $o_i$, with preceding context $H_{i,m}$, we define the importance ratio as
\begin{equation}
\rho_{i,m}(\theta)=
\frac{\pi_\theta(o_{i,m}\mid H_{i,m})}
{\pi_{\theta_{\mathrm{old}}}(o_{i,m}\mid H_{i,m})}.
\label{eq:importance_ratio}
\end{equation}
Using $A_i$ from Eq.~\ref{eq:combined_advantage}, RA-GRPO maximizes
\begin{equation}
\begin{aligned}
&\mathcal J_{\mathrm{RA\text{-}GRPO}}(\theta)=\\[2pt]
&\mathbb E_{x,\{o_i\},i,m}\!\left[
\min\!\left(\rho_{i,m}A_i,
\operatorname{clip}(\rho_{i,m},1-\varepsilon_c,1+\varepsilon_c)A_i\right)
\right].
\end{aligned}
\label{eq:ra_grpo_objective}
\end{equation}

%% file: sec/table_main_results.tex
\providecommand{\accalls}[2]{#1\,{\color{black!65}/\,#2}}
\providecommand{\best}[1]{\textbf{#1}}
\providecommand{\second}[1]{\underline{#1}}

\begin{table*}[t]
\centering
\caption{\textbf{RA-GRPO reduces mean tool calls by 15.38\% relative to the tool-enabled backbone and by 12.39\% relative to Vanilla GRPO at comparable accuracy.} Entries report answer accuracy in percent / mean tool calls per question. Macro averages weight the seven benchmarks equally. $\Delta$ denotes RA-GRPO minus tool-enabled Qwen3-VL-8B-Instruct, with accuracy differences in percentage points and call differences in calls per question. Bold and underlined values indicate the highest and second-highest accuracies in each column.}
\label{tab:main_results}
\begingroup
\footnotesize
\setlength{\tabcolsep}{0.8pt}
\renewcommand{\arraystretch}{1.18}
\begin{tabular*}{\textwidth}{@{\extracolsep{\fill}}l*{7}{c}cc@{}}
\toprule
& \multicolumn{7}{c}{\textbf{Benchmark Acc. $\uparrow$ / Calls}}
& \multicolumn{2}{c}{\textbf{Macro Avg.}} \\
\cmidrule(lr){2-8}
\cmidrule(l){9-10}
\textbf{Method} & HR-4K & HR-8K & V$^\star$Bench & VisualProbe & ZoomBench & MMStar & MME-RW-Lite & Acc. $\uparrow$ & Calls \\
\midrule
\multicolumn{10}{@{}l}{\emph{Backbones}} \\
Qwen2.5-VL-7B-Instruct
& \accalls{69.75}{0.00} & \accalls{63.88}{0.00} & \accalls{79.58}{0.00} & \accalls{6.99}{0.00} & \accalls{45.85}{0.00} & \accalls{56.53}{0.00} & \accalls{48.31}{0.00} & 52.98 & 0.00 \\
Qwen3-VL-8B-Instruct w/o tool
& \accalls{73.50}{0.00} & \accalls{70.50}{0.00} & \accalls{74.87}{0.00} & \accalls{27.18}{0.00} & \accalls{38.15}{0.00} & \accalls{66.27}{0.00} & \accalls{49.61}{0.00} & 57.15 & 0.00 \\
\addlinespace[2pt]
\midrule
\multicolumn{10}{@{}l}{\emph{Visual tool-use models}} \\
Pixel-Reasoner~\cite{pixelreasoner}
& \accalls{73.88}{0.86} & \accalls{67.75}{0.84} & \accalls{82.72}{0.50} & \accalls{33.79}{0.90} & \accalls{46.80}{0.90} & \accalls{62.27}{0.66} & \accalls{53.05}{0.90} & 60.04 & 0.79 \\
CodeV-RL~\cite{codev}
& \accalls{74.00}{1.04} & \accalls{69.50}{1.28} & \accalls{83.25}{1.74} & \accalls{17.86}{1.96} & \accalls{48.58}{0.54} & \accalls{27.67}{0.06} & \accalls{49.50}{1.35} & 52.91 & 1.14 \\
Thyme-RL~\cite{thyme}
& \accalls{77.62}{0.13} & \accalls{71.50}{0.12} & \accalls{83.77}{0.12} & \accalls{28.35}{0.08} & \accalls{45.50}{0.22} & \accalls{62.33}{0.01} & \accalls{\second{54.77}}{0.07} & 60.55 & 0.11 \\
Mini-o3-7B~\cite{minio3}
& \accalls{70.12}{2.49} & \accalls{66.00}{2.80} & \accalls{85.34}{2.20} & \accalls{34.76}{4.68} & \accalls{45.02}{2.97} & \accalls{54.93}{1.37} & \accalls{42.26}{4.81} & 56.92 & 3.05 \\
\addlinespace[2pt]
\midrule
Qwen3-VL-8B-Instruct w/ tool
& \accalls{81.38}{1.43} & \accalls{76.38}{1.50} & \accalls{84.29}{0.90} & \accalls{38.64}{1.52} & \accalls{50.95}{1.15} & \accalls{64.13}{0.75} & \accalls{47.42}{0.93} & 63.31 & 1.17 \\
$+$ Vanilla GRPO
& \accalls{\best{82.88}}{1.17} & \accalls{\best{80.62}}{1.16} & \accalls{\best{92.15}}{1.07} & \accalls{\second{41.36}}{1.15} & \accalls{\second{56.64}}{1.04} & \accalls{\second{67.47}}{1.25} & \accalls{\second{54.77}}{1.07} & \second{67.98} & 1.13 \\
\textbf{$+$ RA-GRPO (Ours)}
& \accalls{\second{82.50}}{1.03} & \accalls{\second{78.00}}{1.03} & \accalls{\second{89.01}}{1.00} & \accalls{\best{43.88}}{1.00} & \accalls{\best{58.41}}{1.00} & \accalls{\best{69.27}}{0.90} & \accalls{\best{55.86}}{0.98} & \best{68.13} & 0.99 \\
\addlinespace[2pt]
$\Delta$ Ours $-$ Baseline
& \textcolor{green!50!black}{$+1.12$}\kern0.4pt {\color{black}/\kern0.4pt $-0.40$} & \textcolor{green!50!black}{$+1.62$}\kern0.4pt {\color{black}/\kern0.4pt $-0.47$} & \textcolor{green!50!black}{$+4.72$}\kern0.4pt {\color{black}/\kern0.4pt $+0.10$} & \textcolor{green!50!black}{$+5.24$}\kern0.4pt {\color{black}/\kern0.4pt $-0.52$} & \textcolor{green!50!black}{$+7.46$}\kern0.4pt {\color{black}/\kern0.4pt $-0.15$} & \textcolor{green!50!black}{$+5.14$}\kern0.4pt {\color{black}/\kern0.4pt $+0.15$} & \textcolor{green!50!black}{$+8.44$}\kern0.4pt {\color{black}/\kern0.4pt $+0.05$} & \textcolor{green!50!black}{$+4.82$} & {\color{black}$-0.18$} \\
\bottomrule
\end{tabular*}
\endgroup
\end{table*}

%% file: sec/4_experiments.tex
\begin{figure*}[t]
\centering
\includegraphics[width=\textwidth]{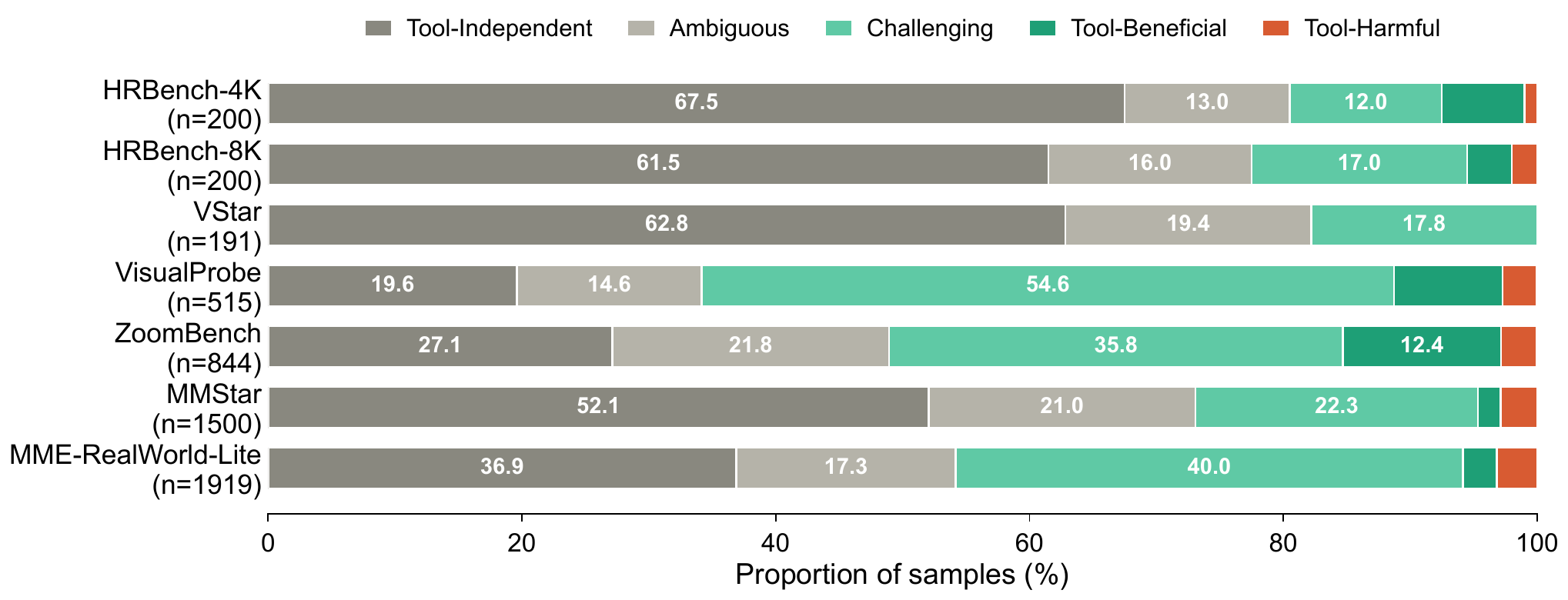}
\caption{\textbf{CAT reveals different tool-utility distributions across the seven benchmarks.} Labels are derived from Qwen3-VL-8B-Instruct rollouts with and without tool access. Bars show the percentage of questions in each category, and $n$ denotes the number of questions. Each HRBench resolution contains 200 original questions; all four option permutations share the voted label for that question. VisualProbe combines its Easy, Medium, and Hard splits.}
\label{fig:cat_distribution}
\end{figure*}

\section{Experiments}
\label{sec:experiments}

\subsection{Experimental Setup}
\label{sec:experimental_setup}

\paragraph{Training Data.}
We construct the reinforcement learning training set from VisualProbe-Train~\cite{minio3} and ZwZ-RL-VQA-10K~\cite{zooming_without_zooming}. After deduplication and solvability filtering, we obtain approximately 10K candidate samples. We retain Tool-Independent, Tool-Beneficial, and Challenging samples and apply additional exploration-quality and consistency filters, yielding approximately 3.3K training examples.

\paragraph{Evaluation Protocol.}
We evaluate visual reasoning and tool-use efficiency on seven benchmarks, namely HRBench-4K, HRBench-8K~\cite{hrbench}, V$^\star$Bench~\cite{vstar}, VisualProbe~\cite{minio3}, ZoomBench~\cite{zooming_without_zooming}, MMStar~\cite{mmstar}, and MME-RealWorld-Lite~\cite{mme_realworld}. VisualProbe results aggregate its Easy, Medium, and Hard splits. We report answer accuracy (Acc.) and mean tool calls per question, denoted as Calls, including failed attempts. Baselines include Qwen2.5-VL-7B-Instruct~\cite{qwen25vl}, Qwen3-VL-8B-Instruct~\cite{qwen3vl} with and without tools, and Vanilla GRPO, which is trained from the same backbone under the same data and training configuration as RA-GRPO, with the efficiency terms removed. We also evaluate open-source models under the same protocol, namely Mini-o3-7B~\cite{minio3}, CodeV-RL~\cite{codev}, Pixel-Reasoner~\cite{pixelreasoner}, and Thyme-RL~\cite{thyme}. We additionally report category-wise accuracy and mean tool calls using shared CAT labels across models.

\paragraph{Implementation Details.}
We train Qwen3-VL-8B-Instruct with RA-GRPO. Training runs for 406 steps, with 8 questions per step, 8 trajectories per question, and at most 4 tool calls per trajectory. Training uses 8 H100 GPUs. Additional details on data preparation, training and generation settings, answer scoring, and VFD implementation are provided in Appendix~\ref{sec:experimental_setup_details}.

\input{sec/table_category_results}
\input{sec/table_ablation_category}

\subsection{Benchmark-wise Results}
\label{sec:main_results}

\paragraph{RA-GRPO matches benchmark accuracy with fewer tool calls.}
Compared with the tool-enabled Qwen3-VL-8B-Instruct backbone, RA-GRPO raises benchmark macro-average accuracy from 63.31\% to 68.13\% and reduces mean tool calls from 1.17 to 0.99, a reduction of 15.38\% according to Table~\ref{tab:main_results}. Vanilla GRPO reaches 67.98\%, so most of this accuracy gain comes from reinforcement learning itself rather than from the efficiency terms. At the benchmark level, RA-GRPO contributes a 12.39\% reduction in calls at comparable accuracy.

\paragraph{RA-GRPO improves visual understanding and reasoning performance.}
Our model improves accuracy over the tool-enabled backbone on all five perception benchmarks and both general benchmarks, and outperforms open-source visual tool-use models. Despite training on perception tasks, RA-GRPO also improves performance on the general benchmarks, with gains of 5.14 and 8.44 percentage points on MMStar and MME-RealWorld-Lite respectively.

Table~\ref{tab:main_results} also shows substantial variation in tool-call frequency across the evaluated models. Benchmark macro-average accuracy, however, cannot show where these differences arise, because it is dominated by the categories that occupy most of each benchmark. Tool-Independent samples account for 19.61\%--67.50\% of the seven benchmarks and are already answered correctly by most models, whereas Tool-Beneficial and Tool-Harmful samples account for at most 15.28\% of each benchmark. A large accuracy change on the latter can therefore shift the benchmark average by a proportionally smaller margin. We next use CAT to report accuracy and tool-call frequency by sample-level tool utility.

\subsection{Category-wise Results}
\label{sec:category_results}

We apply CAT to the seven benchmarks in Table~\ref{tab:main_results}, giving the categorization shown in Figure~\ref{fig:cat_distribution}. Based on this categorization, we report category-wise accuracy and mean tool calls in Table~\ref{tab:category_results}.

\paragraph{CAT reveals the tool-utility distribution of each dataset.}
Tool-utility distributions vary substantially across benchmarks. Under the Qwen3-VL-8B-Instruct teacher and the inference protocol of Appendix~\ref{sec:experimental_setup_details}, Tool-Independent samples account for 67.50\%, 61.50\%, and 62.83\% of HRBench-4K, HRBench-8K, and V$^\star$Bench, respectively, whereas samples meeting the Tool-Beneficial criteria constitute at most 12.44\% of each benchmark. This indicates that even benchmarks emphasizing high-resolution visual understanding contain a substantial proportion of samples that can be reliably answered without tools, so overall accuracy is a poor measure of the ability of a model to solve problems with tools. VisualProbe and ZoomBench, by contrast, specifically select small targets and difficult questions during construction. Consistent with this focus, Challenging samples account for 54.6\% and 35.8\%, respectively, yet both benchmarks also contain Tool-Independent samples. This suggests that prior difficulty-based filtering does not necessarily capture sample-level tool utility and CAT explicitly distinguishes these cases.

\paragraph{RA-GRPO reduces calls in every category and gains most on Tool-Harmful samples.}
Relative to Vanilla GRPO, RA-GRPO reduces mean tool calls by 11.82\%--25.19\% across the five categories and raises the category macro-average accuracy from 59.64\% to 62.99\%. On Tool-Beneficial samples, accuracy increases by 1.30 percentage points while calls fall by 12.28\%, indicating that the reduction removes low-value rather than necessary calls. The largest gain occurs on Tool-Harmful samples, where accuracy rises by 13.86 percentage points and calls fall by 25.19\%. Tool-Independent accuracy is virtually unchanged, differing by $-0.23$ percentage points. Because Tool-Beneficial and Tool-Harmful samples constitute at most 15.28\% of each benchmark, these category-level differences can shift the benchmark macro-average by a proportionally smaller margin in Table~\ref{tab:main_results}, which is why evaluation by tool utility is needed to observe them.

\paragraph{CAT reveals tool-use differences obscured by benchmark scores.}
On Tool-Independent samples, accuracy ranges from 73.07\% to 97.98\% despite widely varying call frequencies, showing that frequent tool use does not necessarily imply better performance. The tool-enabled Qwen3-VL-8B-Instruct backbone achieves a substantially higher macro-average accuracy than Qwen2.5-VL-7B-Instruct in Table~\ref{tab:main_results}, yet scores only 36.14\% on the Tool-Harmful samples, the lowest among the evaluated models. Thus, stronger aggregate performance can coexist with weaknesses on samples susceptible to tool-induced harm. Among open-source methods based on Qwen2.5-VL-7B-Instruct, Pixel-Reasoner and Thyme obtain their largest category-wise accuracy gains in the Tool-Beneficial category after acquiring strong tool-use capabilities, highlighting an improvement pattern hidden by aggregate scores in Table~\ref{tab:main_results}. By contrast, Mini-o3 averages 3.38 calls across categories but improves category macro-average accuracy by only 2.45 percentage points over Qwen2.5-VL-7B-Instruct, suggesting that more frequent interaction does not necessarily yield commensurate gains. CAT therefore distinguishes category-specific strengths and weaknesses that benchmark-level accuracy alone does not reveal.
\input{sec/4_ablation}

%% file: sec/table_category_results.tex
\begin{table*}[t]
\centering
\caption{\textbf{RA-GRPO reduces mean tool calls in all five CAT categories relative to Vanilla GRPO, with the largest accuracy gain on Tool-Harmful samples.} All models are evaluated using shared CAT labels. Acc. denotes answer accuracy in percent; Calls denotes mean tool calls per question, including failed attempts. Cat. Macro Avg. weights the five categories equally. $\Delta$ denotes RA-GRPO minus Vanilla GRPO, with accuracy differences in percentage points and call differences in calls per question. Bold and underlined values indicate the highest and second-highest accuracies in each column.}
\label{tab:category_results}
\begingroup
\footnotesize
\setlength{\tabcolsep}{2.5pt}
\renewcommand{\arraystretch}{1.15}
\begin{tabular*}{\textwidth}{@{\extracolsep{\fill}}l*{6}{w{r}{25pt}w{r}{25pt}}@{}}
\toprule
& \multicolumn{2}{c}{\makebox[0pt]{\textbf{Tool-Independent}}}
& \multicolumn{2}{c}{\makebox[0pt]{\textbf{Tool-Beneficial}}}
& \multicolumn{2}{c}{\makebox[0pt]{\textbf{Challenging}}}
& \multicolumn{2}{c}{\makebox[0pt]{\textbf{Tool-Harmful}}}
& \multicolumn{2}{c}{\makebox[0pt]{\textbf{Ambiguous}}}
& \multicolumn{2}{c}{\makebox[0pt]{\textbf{Cat. Macro Avg.}}} \\
\cmidrule(lr){2-3}
\cmidrule(lr){4-5}
\cmidrule(lr){6-7}
\cmidrule(lr){8-9}
\cmidrule(lr){10-11}
\cmidrule(l){12-13}
\textbf{Method}
& Acc.$\uparrow$ & Calls
& Acc.$\uparrow$ & Calls
& Acc.$\uparrow$ & Calls
& Acc.$\uparrow$ & Calls
& Acc.$\uparrow$ & Calls
& Acc.$\uparrow$ & Calls \\
\midrule
\multicolumn{13}{@{}l}{\emph{Backbones}} \\
Qwen2.5-VL-7B-Instruct
& 81.22 & 0.00 & 36.16 & 0.00 & 15.95 & 0.00 & 53.01 & 0.00 & 42.21 & 0.00 & 45.71 & 0.00 \\
Qwen3-VL-8B-Instruct w/o tool
& \textbf{97.98} & 0.00 & 12.05 & 0.00 & 4.00 & 0.00 & \textbf{88.55} & 0.00 & 45.11 & 0.00 & 49.54 & 0.00 \\
\midrule
\multicolumn{13}{@{}l}{\emph{Thinking-with-images Models}} \\
Pixel-Reasoner
& 86.17 & 0.83 & 49.51 & 0.92 & 20.87 & 0.84 & \underline{64.46} & 0.71 & 49.87 & 0.77 & 54.18 & 0.81 \\
CodeV-RL
& 73.07 & 0.90 & 52.12 & 1.18 & 15.44 & 1.03 & 52.41 & 0.99 & 38.72 & 0.94 & 46.35 & 1.01 \\
Thyme-RL
& 86.30 & 0.09 & 60.26 & 0.18 & \textbf{23.23} & 0.09 & 51.20 & 0.13 & 49.53 & 0.08 & 54.10 & 0.11 \\
Mini-o3-7B
& 79.77 & 2.10 & 49.51 & 3.77 & 18.36 & 4.58 & 52.41 & 3.02 & 40.77 & 3.41 & 48.16 & 3.38 \\
\midrule
Qwen3-VL-8B-Instruct w/ tool
& \underline{95.49} & 0.97 & \textbf{83.06} & 1.34 & 9.13 & 1.08 & 36.14 & 2.24 & 50.55 & 1.19 & 54.87 & 1.36 \\
$+$ Vanilla GRPO
& 95.32 & 1.10 & 75.90 & 1.14 & 20.77 & 1.14 & 47.59 & 1.35 & \underline{58.64} & 1.18 & \underline{59.64} & 1.18 \\
\textbf{$+$ RA-GRPO (Ours)}
& 95.09 & 0.97 & \underline{77.20} & 1.00 & \underline{21.38} & 1.00 & 61.45 & 1.01 & \textbf{59.83} & 0.96 & \textbf{62.99} & 0.99 \\
\addlinespace[2pt]
$\Delta$ Ours $-$ Vanilla
& \textcolor[HTML]{B22222}{$-0.23$} & {\color{black}$-0.13$}
& \textcolor{green!50!black}{$+1.30$} & {\color{black}$-0.14$}
& \textcolor{green!50!black}{$+0.61$} & {\color{black}$-0.14$}
& \textcolor{green!50!black}{$+13.86$} & {\color{black}$-0.34$}
& \textcolor{green!50!black}{$+1.19$} & {\color{black}$-0.22$}
& \textcolor{green!50!black}{$+3.35$} & {\color{black}$-0.19$} \\
\bottomrule
\end{tabular*}
\endgroup
\end{table*}

%% file: sec/table_ablation_category.tex
\begin{table*}[t]
\centering
\caption{\textbf{The full RA-GRPO design achieves the highest category-macro accuracy among the evaluated variants.} VFD $\rightarrow$ All-Redundant treats every eligible call as redundant, while w/o VFD Cost removes the VFD redundancy cost. Uniform Base Cost applies the base cost to all categories, while w/o Base Cost retains only the VFD redundancy cost. Acc. denotes answer accuracy in percent; Calls denotes mean tool calls per question. Cat. Avg. weights the five categories equally. Bold and underlined values indicate the highest and second-highest accuracies in each column.}
\label{tab:ablation_category}

\begingroup
\footnotesize
\setlength{\tabcolsep}{3pt}
\renewcommand{\arraystretch}{1.18}

\begin{tabular*}{\textwidth}{
  @{\extracolsep{\fill}}l*{6}{rr}@{}
}
\toprule
& \multicolumn{2}{c}{
    \textbf{\shortstack{Tool-\\Independent}}}
& \multicolumn{2}{c}{
    \textbf{\shortstack{Tool-\\Beneficial}}}
& \multicolumn{2}{c}{\textbf{Challenging}}
& \multicolumn{2}{c}{
    \textbf{\shortstack{Tool-\\Harmful}}}
& \multicolumn{2}{c}{\textbf{Ambiguous}}
& \multicolumn{2}{c}{\textbf{Cat. Avg.}} \\
\cmidrule(lr){2-3}
\cmidrule(lr){4-5}
\cmidrule(lr){6-7}
\cmidrule(lr){8-9}
\cmidrule(lr){10-11}
\cmidrule(l){12-13}

\textbf{Variant}
& Acc.$\uparrow$ & Calls
& Acc.$\uparrow$ & Calls
& Acc.$\uparrow$ & Calls
& Acc.$\uparrow$ & Calls
& Acc.$\uparrow$ & Calls
& Acc.$\uparrow$ & Calls \\
\midrule

Vanilla GRPO
& \textbf{95.32} & 1.10 & 75.90 & 1.14 & \underline{20.77} & 1.14 & 47.59 & 1.35 & \underline{58.64} & 1.18 & 59.64 & 1.18 \\

\midrule

VFD $\rightarrow$ All-Redundant
& 95.19 & 0.27
& 70.68 & 0.74
& 17.18 & 0.55
& \textbf{66.87} & 0.64
& 54.21 & 0.51
& \underline{60.83} & 0.54 \\

w/o VFD Cost
& \underline{95.22} & 1.20
& \textbf{78.83} & 1.25
& 17.79 & 1.23
& 50.00 & 1.61
& 57.70 & 1.34
& 59.91 & 1.33 \\

\addlinespace[3pt]

Uniform Base Cost
& 93.17 & 1.07
& \underline{78.18} & 1.06
& 17.18 & 1.12
& 50.00 & 1.25
& 54.81 & 1.18
& 58.67 & 1.14 \\

w/o Base Cost with $K=0$
& 93.67 & 1.20
& 74.59 & 1.23
& \underline{20.77} & 1.44
& 51.81 & 1.81
& 57.62 & 1.51
& 59.69 & 1.44 \\

\midrule

\textbf{RA-GRPO (Ours)}
& 95.09 & 0.97
& 77.20 & 1.00
& \textbf{21.38} & 1.00
& \underline{61.45} & 1.01
& \textbf{59.83} & 0.96
& \textbf{62.99} & 0.99 \\

\bottomrule
\end{tabular*}

\par\smallskip

\endgroup
\vspace{-6pt}
\end{table*}

%% file: sec/4_ablation.tex
\raggedbottom
\subsection{Ablation Study}
\label{sec:ablation}

To better understand the contributions of VFD redundancy estimation and the category-conditioned base cost, we remove or replace the VFD redundancy cost and vary the scope of the base cost. Table~\ref{tab:ablation_category} reports accuracy and tool-call counts for each variant across the five CAT categories. Table~\ref{tab:ablation_variants} reports benchmark and category macro-average accuracy for each variant. Detailed ablation results are provided in Appendix~\ref{sec:ablation_details}.

\paragraph{Effect of VFD redundancy estimation.}
Estimating redundancy from feedback dependency gives a better balance between accuracy and call reduction than either variant, reaching the highest category macro-average accuracy of 62.99\% in Table~\ref{tab:ablation_category}. Replacing VFD with All-Redundant, which treats every eligible call as redundant, cuts calls further in every category and achieves the highest Tool-Harmful accuracy of 66.87\%, consistent with the view that samples harmed by tools benefit from fewer calls. Its Tool-Beneficial accuracy, however, drops to 70.68\%, showing that indiscriminate call reduction can reduce accuracy on Tool-Beneficial samples. Removing the VFD cost has the opposite effect. Category-macro calls rise to 1.33, and accuracy falls on Challenging, Tool-Harmful, and Ambiguous samples, with Tool-Beneficial accuracy 1.63 points above RA-GRPO obtained with 25\% more calls. These comparisons support using feedback dependency to decide which calls to discourage, rather than discouraging all of them or none.

\begin{table}[tb]
\centering
\caption{\textbf{VFD-based costing and the category-conditioned base cost each contribute to the accuracy--call balance.} VFD $\rightarrow$ All-Redundant treats every eligible call as redundant, while w/o VFD Cost removes the VFD redundancy cost. Uniform Base Cost applies the base cost to all categories, while w/o Base Cost retains only the VFD redundancy cost. Bench. Avg. and Cat. Avg. report accuracy in percent, averaged equally over the seven benchmarks and five CAT categories, respectively. $\Delta$ Cat. denotes the difference between each variant and Vanilla GRPO in percentage points. Bold and underlined values indicate the highest and second-highest accuracies.}
\label{tab:ablation_variants}
\begingroup
\small
\setlength{\tabcolsep}{3pt}
\renewcommand{\arraystretch}{1.18}
\begin{tabular*}{\columnwidth}{@{\extracolsep{\fill}}lrrr@{}}
\toprule
\textbf{Variant}
& \textbf{\shortstack{Bench.\\Avg.}}
& \textbf{\shortstack{Cat.\\Avg.}}
& $\boldsymbol{\Delta}$ \textbf{Cat.} \\
\midrule
Vanilla GRPO & \underline{67.98} & 59.64 & $\phantom{+}0.00$ \\
\midrule
VFD $\rightarrow$ All-Redundant & 66.53 & \underline{60.83} & $+1.19$ \\
w/o VFD Cost & 67.14 & 59.91 & $+0.27$ \\
Uniform Base Cost & 65.07 & 58.67 & $-0.97$ \\
w/o Base Cost with $K=0$ & 67.40 & 59.69 & $+0.05$ \\
\midrule
\textbf{RA-GRPO (Ours)} & \textbf{68.13} & \textbf{62.99} & $+3.35$ \\
\bottomrule
\end{tabular*}
\endgroup
\vspace{-14pt}
\end{table}

\paragraph{Effect of the Category-conditioned Base Cost.}
The category-conditioned base cost improves tool-use efficiency. The full RA-GRPO method achieves higher benchmark macro-average accuracy than both base-cost variants in Table~\ref{tab:ablation_variants}. Compared with w/o Base Cost with $K=0$, which uses only the VFD redundancy cost, RA-GRPO achieves higher accuracy and fewer calls in all five CAT categories. Call reductions are at least 18.70\%, with the smallest reduction occurring on Tool-Beneficial samples in Table~\ref{tab:ablation_category}. The Uniform Base Cost variant applies the base cost to all categories without distinguishing their tool utility, and it yields lower benchmark and category macro-average accuracy and more calls than RA-GRPO. These results support targeting the base cost at Tool-Independent samples while using VFD to assess redundancy in categories, and a broader call penalty does not necessarily produce fewer calls.

%% file: sec/5_conclusion.tex
\section{Conclusion}
\label{sec:conclusion}

We introduced $R^2$, a framework connecting sample-level tool-utility analysis, turn-level feedback diagnosis, and redundancy-aware policy optimization. CAT reveals category-specific strengths and weaknesses obscured by aggregate benchmark scores, while VFD measures the dependence of subsequent reasoning on visual feedback without an external evaluator. RA-GRPO combines this signal with a category-conditioned base call cost to assign relative efficiency advantages among correct trajectories. Across seven benchmarks, RA-GRPO achieves accuracy comparable to Vanilla GRPO under matched training conditions while reducing mean tool calls by 12.39\%. Category-wise evaluation further shows that RA-GRPO improves category macro-average accuracy by 3.35 points and reduces calls in every category, with its largest accuracy gain on Tool-Harmful samples. These findings support evaluating and optimizing visual tool use in terms of sample-level tool utility, rather than by answer accuracy or call frequency alone.

\paragraph{Limitations and Future Work.}
\looseness=-1
Although Appendix~\ref{sec:vfd_other_tools} presents case studies of VFD on other visual tools, our experiments focus on crop-and-zoom tools and do not investigate training models to use multiple tools. CAT labels can reflect the tool-utility distribution of a dataset, but the results still vary with the teacher model and inference protocol. VFD serves as a proxy for feedback dependency rather than a direct measure of the causal contribution of a call, and its estimates depend on the sampled calls and alternative observations. Moreover, reducing repeated observations does not establish that a model can reliably decide whether to invoke a tool in the first place. Future work will investigate more robust dependency estimation and extend the framework to additional visual tools and model families. We also encourage future benchmarks and training datasets to account for the distribution of sample-level tool utility.

%% file: sec/llm_usage.tex
\section{LLM Usage Statement}
\label{sec:llm_usage}

We used ChatGPT during manuscript preparation to assist with language translation and editing, text organization, LaTeX typesetting, verification of bibliographic information, and improvements to the presentation of figures and tables. Model outputs served as references and were subject to review and verification by the authors. The authors take full responsibility for the accuracy and completeness of the final manuscript, as well as the experimental results and research conclusions.

%% file: sec/X_vfd_validation.tex
\section{Offline Validation of Visual Feedback Dependency}
\label{sec:vfd_offline_validation}

\subsection{Evaluation Protocol}

We evaluate whether Visual Feedback Dependency (VFD) distinguishes necessary from redundant visual tool calls. We collect trajectories generated by Qwen3-VL-8B-Instruct with access to crop-and-zoom tools on VisualProbe and ZoomBench. Each evaluation unit comprises a tool call, its returned crop, and the model response immediately following it. A question may contribute multiple calls.

We use \texttt{gemini-3-flash-preview} as the model identifier to obtain annotations. The annotator assigns one of four necessity labels, namely \emph{necessary}, \emph{possibly redundant}, \emph{redundant}, or \emph{unclear}.

We compute VFD for each reasoning window by comparing its log-likelihood under the actual visual feedback with the mean value under three alternative visual observations. We report ROC-AUC under two label mappings. The \textbf{strict comparison} includes only necessary and redundant calls, whereas the \textbf{relaxed comparison} combines possibly redundant and redundant calls into the negative class. Unclear annotations are excluded. Necessary calls form the positive class, so a higher AUC indicates a stronger ability to rank necessary calls above redundant ones. Statistical comparisons use question-level clustering to account for multiple windows from the same question. 

\subsection{Offline Discrimination and Redundancy Enrichment}
\label{sec:vfd_offline_results}

We evaluate offline whether VFD provides additional information for identifying redundant tool calls. Table~\ref{tab:vfd_offline_auc} reports results using noise and random crops as alternative visual feedback, while Table~\ref{tab:vfd_offline_enrichment} reports redundancy enrichment among calls with low VFD scores.

\par\addvspace{\dimexpr\intextsep\relax}\noindent\begin{minipage}{\columnwidth}
\captionsetup{type=table}
\caption{ROC-AUC using noise and random crops as alternative visual feedback. Windows denotes the number of tool-call turns included in the analysis, each paired with its subsequent reasoning window. The strict comparison additionally excludes possibly redundant calls.}
\label{tab:vfd_offline_auc}
\centering\footnotesize
\setlength{\tabcolsep}{2pt}\renewcommand{\arraystretch}{1.15}
\begin{tabular*}{\linewidth}{@{\extracolsep{\fill}}lrcccc@{}}
\toprule
& & \multicolumn{2}{c}{Strict} & \multicolumn{2}{c}{Relaxed} \\
\cmidrule(lr){3-4}\cmidrule(l){5-6}
Dataset & Windows & Noise & Crop & Noise & Crop \\
\midrule
VisualProbe & 836 & 0.760 & 0.766 & 0.750 & 0.748 \\
ZoomBench & 919 & 0.652 & 0.638 & 0.616 & 0.608 \\
\bottomrule
\end{tabular*}
\end{minipage}\par\addvspace{\dimexpr\intextsep\relax}

The analysis in Table~\ref{tab:vfd_offline_auc} includes 836 tool-call turns from VisualProbe and 919 from ZoomBench, each annotated by a VLM for tool-use necessity. Under the strict comparison, random-crop feedback yields ROC-AUC values of 0.766 on VisualProbe and 0.638 on ZoomBench. Noise feedback gives similar values of 0.760 and 0.652, respectively.

\par\addvspace{\dimexpr\intextsep\relax}\noindent\begin{minipage}{\columnwidth}
\captionsetup{type=table}
\caption{Redundancy enrichment among windows with the lowest 20\% of VFD scores computed on reasoning text using noise as alternative visual feedback. Strict redundancy counts only calls labeled \emph{redundant}.}
\label{tab:vfd_offline_enrichment}
\centering\footnotesize
\setlength{\tabcolsep}{2pt}\renewcommand{\arraystretch}{1.15}
\begin{tabular*}{\linewidth}{@{\extracolsep{\fill}}lrrr@{}}
\toprule
Dataset & \shortstack{Selected\\redundancy} & \shortstack{Overall\\redundancy} & \shortstack{Enrich-\\ment} \\
\midrule
VisualProbe & 61.99\% & 28.76\% & 2.16$\times$ \\
ZoomBench & 19.15\% & 13.97\% & 1.37$\times$ \\
\bottomrule
\end{tabular*}
\end{minipage}\par\addvspace{\dimexpr\intextsep\relax}

Table~\ref{tab:vfd_offline_enrichment} shows that selecting calls with low VFD scores yields 2.16-fold and 1.37-fold enrichment of annotated redundant calls on VisualProbe and ZoomBench, respectively. Low VFD scores therefore identify subsets with higher concentrations of annotated redundant calls on both datasets, supporting their use as a signal of potential redundancy.

\subsection{Agreement with Human Annotations}

We assess annotation reliability on 93 ZoomBench windows from trajectories generated by Qwen3-VL-8B-Instruct with tool access, each with paired human and VLM judgments. Under the relaxed comparison, agreement between human and VLM annotations reaches 74.19\%, with a binary macro-F1 of 0.728.

With human labels on the same review cohort, VFD achieves ROC-AUC values of 0.807 and 0.728 under the strict and relaxed comparisons, respectively. This provides supplementary evidence that its ability to distinguish necessary from redundant calls is not solely an artifact of VLM annotations. 

\subsection{VLM Annotation Inputs and Prompt}

The annotator receives an anonymized sample identifier, the question, the original image, visible reasoning and crops preceding the current call, the current tool request, the returned crop, and the immediately following response. We exclude subsequent turns, reference answers, correctness labels, model identity, VFD scores, human annotations, and CAT labels.

The annotator uses temperature 0 and a maximum output budget of 8,192 tokens. API or parsing failures result in missing annotations and receive no necessity label.

\textbf{Annotation system prompt.}

\begingroup\small\ttfamily\raggedright
\setlength{\parskip}{0.5\baselineskip}
You independently annotate one visual question-answering model's current zoom-tool turn. Use only the supplied question, original image, visible earlier history, current tool request, current crop, and immediate response. The assistant can give a plausible or even correct answer from earlier evidence or a guess; that does not prove the crop helped. A useful crop can be redundant for a simple question. Do not use future actions, reference answer, model identity or hidden outcome.

**Necessity asks whether this particular crop was needed to obtain relevant new visual evidence.** Compare it against what a careful reader could reasonably perceive from the original image and earlier crops at the provided resolution. `necessary`: the crop makes a question-relevant detail distinguishable that was difficult to resolve before (tiny text/object, low contrast, similar items, inaccessible region). The response need not be correct, but there should be evidence the crop could genuinely provide. `redundant`: the decisive content was already clear, or this crop adds no relevant detail. `possibly\_redundant`: the crop plausibly confirms or clarifies a detail but the original/history may have sufficed. `unclear`: insufficient evidence. Do not call a crop necessary merely because the model called the tool or pointed at the correct region.

Return exactly one JSON object with exactly these keys: `sample\_id`, `necessity`, `confidence`, `evidence`. Copy the supplied Sample ID verbatim into `sample\_id`. Choose `confidence` from `high`, `medium`, `low`. In `evidence`, cite visible features and immediate-response claims in at most two sentences. No Markdown or code fence.
\par\endgroup

%% file: sec/X_cat_details.tex
\section{CAT Analysis Details}
\label{sec:cat_details}

\subsection{CAT Classification Rules}
\label{sec:cat_classification_rules}

\paragraph{Utility classification.}
For each question, we generate five tool-enabled and five tool-disabled rollouts. Let $k_{\mathrm{T}}$ and $k_{\mathrm{D}}$ denote the numbers of correct answers with and without tool access, respectively, and let $k_{\mathrm{C}}$ denote the number of tool-enabled rollouts that invoke at least one tool. The classification rules in Table~\ref{tab:cat_rules} are applied from top to bottom.

\par\addvspace{\dimexpr\intextsep\relax}\noindent\begin{minipage}{\columnwidth}
\captionsetup{type=table}
\caption{\textbf{CAT classification rules using five tool-enabled and five tool-disabled rollouts per question.} $k_{\mathrm T}$ and $k_{\mathrm D}$ count correct answers in the two settings; $k_{\mathrm C}$ counts tool-enabled rollouts with at least one tool call. Rules are applied from top to bottom. A dash indicates no constraint.}
\label{tab:cat_rules}
\centering
\footnotesize
\setlength{\tabcolsep}{4pt}
\renewcommand{\arraystretch}{1.12}
\begin{tabular*}{\linewidth}{@{\extracolsep{\fill}}lccc@{}}
\toprule
Category & $k_{\mathrm{T}}$ & $k_{\mathrm{D}}$ & $k_{\mathrm{C}}$ \\
\midrule
Tool-Independent & $\geq4$ & $\geq4$ & --- \\
Tool-Beneficial & $\geq4$ & $\leq2$ & $\geq3$ \\
Challenging & $\leq1$ & $\leq1$ & --- \\
Tool-Harmful & $\leq2$ & $\geq4$ & $\geq3$ \\
Ambiguous & \multicolumn{3}{c}{All remaining cases} \\
\bottomrule
\end{tabular*}
\end{minipage}\par\addvspace{\dimexpr\intextsep\relax}

\paragraph{Benchmark accounting.}
Category proportions are computed at the question level rather than over rollouts or calls. Each HRBench split contains 200 original questions. We assign each question and its four option permutations the label with the most votes, assigning ties to Ambiguous. VisualProbe's Easy, Medium, and Hard splits form disjoint subsets of the full benchmark and are not counted again in the overall totals.
\begin{table*}[!t]
\centering
\caption{\textbf{Similar CAT distributions obtained with Qwen3-VL-8B-Instruct and Mini-o3 as tool-enabled teachers suggest that CAT can help characterize properties of the datasets themselves.} Entries report the percentage of questions in each category, and $n$ denotes the number of questions. Indented VisualProbe rows provide a breakdown by difficulty split.}
\label{tab:cat_teacher_distributions}
\begingroup
\footnotesize
\setlength{\tabcolsep}{3pt}
\renewcommand{\arraystretch}{1.13}
\begin{tabular*}{\textwidth}{@{\extracolsep{\fill}}lr*{5}{r}@{}}
\toprule
Benchmark & $n$ & Tool-Independent & Tool-Beneficial & Challenging & Tool-Harmful & Ambiguous \\
\midrule
\multicolumn{7}{@{}l}{\textbf{Qwen3-VL-8B-Instruct tool-enabled teacher}} \\
HRBench-4K & 200 & 67.50\% & 6.50\% & 12.00\% & 1.00\% & 13.00\% \\
HRBench-8K & 200 & 61.50\% & 3.50\% & 17.00\% & 2.00\% & 16.00\% \\
V$^\star$Bench & 191 & 62.83\% & 0.00\% & 17.80\% & 0.00\% & 19.37\% \\
VisualProbe overall & 515 & 19.61\% & 8.54\% & 54.56\% & 2.72\% & 14.56\% \\
\quad Easy & 141 & 21.28\% & 12.06\% & 41.84\% & 4.96\% & 19.86\% \\
\quad Medium & 268 & 19.40\% & 6.34\% & 60.82\% & 1.12\% & 12.31\% \\
\quad Hard & 106 & 17.92\% & 9.43\% & 55.66\% & 3.77\% & 13.21\% \\
ZoomBench & 844 & 27.13\% & 12.44\% & 35.78\% & 2.84\% & 21.80\% \\
MMStar & 1,500 & 52.07\% & 1.80\% & 22.27\% & 2.87\% & 21.00\% \\
MME-RealWorld-Lite & 1,919 & 36.89\% & 2.66\% & 39.97\% & 3.18\% & 17.30\% \\
\midrule
\multicolumn{7}{@{}l}{\textbf{Mini-o3 tool-enabled teacher}} \\
HRBench-4K & 200 & 58.50\% & 4.00\% & 12.50\% & 7.50\% & 17.50\% \\
HRBench-8K & 200 & 42.50\% & 1.50\% & 15.00\% & 15.00\% & 26.00\% \\
V$^\star$Bench & 191 & 70.68\% & 12.04\% & 4.71\% & 2.62\% & 9.95\% \\
VisualProbe overall & 515 & 14.17\% & 7.96\% & 49.71\% & 7.96\% & 20.19\% \\
\quad Easy & 141 & 13.48\% & 9.93\% & 36.17\% & 12.77\% & 27.66\% \\
\quad Medium & 268 & 14.55\% & 7.46\% & 55.22\% & 5.97\% & 16.79\% \\
\quad Hard & 106 & 14.15\% & 6.60\% & 53.77\% & 6.60\% & 18.87\% \\
ZoomBench & 844 & 20.85\% & 15.40\% & 31.28\% & 8.89\% & 23.58\% \\
MMStar & 1,500 & 42.00\% & 2.27\% & 19.73\% & 8.80\% & 27.20\% \\
MME-RealWorld-Lite & 1,919 & 28.61\% & 6.41\% & 32.83\% & 11.67\% & 20.48\% \\
\bottomrule
\end{tabular*}
\endgroup
\end{table*}

\subsection{CAT Distributions across Teacher Models}
\label{sec:cat_teacher_comparison}
We compare Qwen3-VL-8B-Instruct and Mini-o3 as tool-enabled teachers, using Qwen3-VL-8B-Instruct without tools as the shared tool-disabled reference. Table~\ref{tab:cat_teacher_distributions} reports the percentage of questions in each category under the classification rules and counting conventions described above.

Across both tool-enabled teachers, Tool-Independent is the largest category on HRBench-4K, HRBench-8K, V$^\star$Bench, and MMStar, whereas Challenging is the largest category on VisualProbe, ZoomBench, and MME-RealWorld-Lite. VisualProbe has the highest proportion of Challenging samples under both teachers, while Tool-Beneficial samples remain a minority across all seven benchmarks. These shared patterns suggest that CAT can help characterize dataset-level tool utility, even though the exact category proportions vary with the tool-enabled teacher.

%% file: sec/X_experimental_setup.tex
\raggedbottom
\section{Experimental Setup Details}
\label{sec:experimental_setup_details}

\subsection{Training and Generation Settings}
\label{sec:training_settings}
Table~\ref{tab:training_settings} lists the training configuration, and Table~\ref{tab:generation_settings} compares the generation settings used during training and evaluation. The base call cost $K$ applies only to Tool-Independent samples.

\begin{table}[!hb]
\captionsetup{justification=raggedright,singlelinecheck=false}
\caption{Training configuration for Vanilla GRPO and RA-GRPO.}
\label{tab:training_settings}
\centering\footnotesize
\setlength{\tabcolsep}{3pt}\renewcommand{\arraystretch}{1.15}
\begin{tabular*}{\linewidth}{@{\extracolsep{\fill}}ll@{}}
\toprule
Parameter & Setting \\
\midrule
Optimizer & AdamW \\
Learning rate & $5\times10^{-7}$ \\
Training examples / steps & 3,254 / 406 \\
Rollouts & 8 \\
KL coefficient & 0 \\
Base call cost $K$ & 0.5 \\
Efficiency weight $\beta$ & 0.2 \\
Tool-call budget $B$ & 4 \\
Hardware & 8 H100 GPUs \\
Training time & $\sim$13 h \\
\bottomrule
\end{tabular*}
\end{table}

\begin{table}[!hb]

\caption{Generation settings for RL training and the main evaluations. Image budgets are expressed in megapixels (MP).}
\label{tab:generation_settings}
\centering\footnotesize
\setlength{\tabcolsep}{2pt}\renewcommand{\arraystretch}{1.15}
\begin{tabular*}{\linewidth}{@{\extracolsep{\fill}}lcc@{}}
\toprule
Setting & RL training & Evaluation \\
\midrule
Temperature & 1.0 & 0\\
Top-$p$ & 1.0 & 1.0 \\
Total response budget in tokens & 8,192 & 8,192 \\
Maximum context in tokens & 20,480 & 131,072 \\
Maximum image budget & 2 MP & 16 MP \\
Maximum tool calls & 4 & 4 \\
\bottomrule
\end{tabular*}
\end{table}

We use the following system prompt during RL stage.
\begin{quote}
\small\ttfamily\raggedright
You are a helpful assistant.

\medskip
Think first, and then decide whether to call an available tool or provide the final answer. If the available information is sufficient to answer the question, answer directly.

\medskip
End your final response with exactly one line in this format:\\
Final answer: <answer>
\end{quote}
The tool \texttt{image\_zoom\_in\_tool} returns a cropped image given a crop box \texttt{bbox\_2d} with coordinates normalized to 0--1000, the index \texttt{img\_idx} of a currently visible image, and an object description \texttt{label}.

We extract answers from the final response. We evaluate multiple-choice answers by matching options and open-ended answers using deterministic text normalization and conservative equivalence rules. Neither procedure uses an external judging model. 

\subsection{Data Preparation and CAT Settings}
\label{sec:data_preparation}
\begin{figure}[!hb]
\centering
\includegraphics[width=\linewidth]{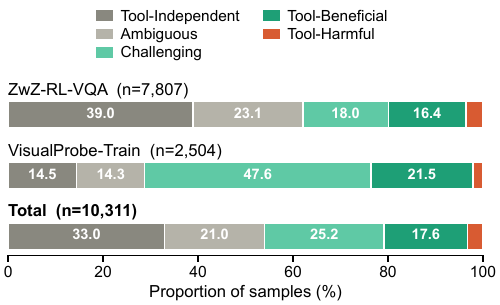}

\caption{\textbf{CAT distributions in the training candidate pool.} Bars show the percentage of candidates in each category for ZwZ-RL-VQA-10K, VisualProbe-Train, and in total.}
\label{fig:training_cat_distribution}
\end{figure}

Figure~\ref{fig:training_cat_distribution} shows the CAT category proportions for each training data source and the overall candidate pool.

We construct the training data from VisualProbe-Train, introduced in Mini-o3~\cite{minio3}, and a sample of 10K examples from ZwZ-RL-VQA-10K~\cite{zooming_without_zooming}. After deduplicating by image and question, we generate 8 tool-enabled rollouts per example with Qwen3-VL-8B-Instruct. We exclude examples for which all 8 answers are incorrect, leaving 10,311 candidates. We assign CAT labels using the classification rules in Table~\ref{tab:cat_rules}, then apply the exploration-quality and consistency filters described below to obtain the final 3,254 training examples.

\paragraph{Exploration statistics.}
For each training candidate, we compute exploration statistics from its five tool-enabled rollouts. We first represent each recorded call with tool arguments by a set of label tokens. Specifically, we lowercase its \texttt{label}, tokenize it using the regular expression \verb|\w+|, and remove duplicate tokens. Let $S_{r,j}$ denote this token set for call $j$ in rollout $r$, and let $L_r$ be the number of recorded call labels in that rollout.

To measure consistency across rollouts, we combine all label tokens within each rollout into a single set,
$P_r=\bigcup_{j=1}^{L_r}S_{r,j}$.
The label-based exploration consistency proxy is the average Jaccard similarity over the ten pairs of rollouts,
\begin{equation}
\rho=\frac{1}{10}
\sum_{1\leq r<s\leq5}
\operatorname{Jaccard}(P_r,P_s).
\label{eq:cat_rho}
\end{equation}
Jaccard similarity is the number of shared tokens divided by the number of tokens in the union. We assign a similarity of $1$ when both sets are empty and $0$ when only one is empty. Thus, $\rho$ measures overlap between rollout-level label sets and serves as a proxy for exploration consistency.

To measure exploration quality within each rollout, we approximately deduplicate its call-label sets in execution order. Starting with an empty collection, we retain a set only if it differs from every previously retained set by more than two tokens, counting tokens present in either set but not both. Let $U_r$ be the number of retained sets. The exploration quality score is
\begin{equation}
Q_r=
\begin{cases}
\dfrac{U_r}{L_r}\,0.9^{L_r-U_r}, & L_r>0,\\[3pt]
0, & L_r=0.
\end{cases}
\label{eq:cat_quality}
\end{equation}
The ratio $U_r/L_r$ measures label diversity, while the factor $0.9^{L_r-U_r}$ penalizes repetition. For each candidate, we compute the mean and population standard deviation across all five rollouts, including those with no calls,
\begin{equation}
\bar Q=\frac{1}{5}\sum_{r=1}^{5}Q_r,\qquad
\sigma_Q=
\sqrt{\frac{1}{5}\sum_{r=1}^{5}(Q_r-\bar Q)^2}.
\label{eq:cat_quality_stats}
\end{equation}

\paragraph{Training-only filtering.}
We apply the additional eligibility conditions in Table~\ref{tab:cat_training_filter} only to training candidates. Tool-Harmful and Ambiguous candidates are excluded. The remaining categories must satisfy $\bar Q\geq0.2$ and $\sigma_Q\leq0.5$. Tool-Beneficial candidates additionally require $\rho\geq0.5$, whereas Challenging candidates require $\rho<0.5$. Tool-Independent candidates have no consistency constraint. We exclude a candidate if any statistic required for its category is unavailable.

\par\addvspace{\dimexpr\intextsep\relax}\noindent\begin{minipage}{\columnwidth}
\captionsetup{type=table}
\caption{\textbf{Additional eligibility conditions for training candidates.} $\bar Q$ and $\sigma_Q$ denote the mean and population standard deviation of exploration quality, and $\rho$ denotes label-based exploration consistency. Candidates must satisfy all conditions for their category and are excluded if any required statistic is unavailable. A dash indicates no consistency constraint.}
\label{tab:cat_training_filter}
\centering
\footnotesize
\setlength{\tabcolsep}{2pt}
\renewcommand{\arraystretch}{1.12}
\begin{tabular*}{\linewidth}{@{\extracolsep{\fill}}lcc@{}}
\toprule
Category & Quality conditions & Consistency \\
\midrule
Tool-Independent & $\bar Q\geq0.2,\ \sigma_Q\leq0.5$ & --- \\
Tool-Beneficial & $\bar Q\geq0.2,\ \sigma_Q\leq0.5$ & $\rho\geq0.5$ \\
Challenging & $\bar Q\geq0.2,\ \sigma_Q\leq0.5$ & $\rho<0.5$ \\
Tool-Harmful & \multicolumn{2}{c}{Excluded} \\
Ambiguous & \multicolumn{2}{c}{Excluded} \\
\bottomrule
\end{tabular*}
\end{minipage}\par\addvspace{\dimexpr\intextsep\relax}

\addtocounter{table}{1}
\input{sec/table_ablation_details}
\addtocounter{table}{-2}

\subsection{VFD Implementation Details}
\label{sec:vfd_implementation}
A call is eligible for VFD scoring if it successfully returns a crop, has a valid, nonempty subsequent text window, and allows the construction of alternative crops of the same size that do not overlap with the returned crop. We prioritize the reasoning text following the feedback. If the model proceeds directly to a final answer without reasoning, an answer window may be used, but answer windows are excluded from threshold initialization. Calls for which alternative crops cannot be constructed are excluded from the eligible set.

We sample one eligible call per trajectory and use three random crops as alternative feedback. During the first 10\% of training steps, we estimate a separate 30th-percentile VFD threshold for each data source using reasoning windows only. We then freeze these thresholds.

During supervision, we set $z_i=1$ when call records are complete but the sampled VFD score is unavailable. An empty eligible set yields zero redundancy cost. We exclude failed calls and calls with incomplete records from this cost comparison.
Table~\ref{tab:vfd_reproduction} summarizes VFD implementation settings.

\par\addvspace{\dimexpr\intextsep\relax}\noindent\begin{minipage}{\columnwidth}
\captionsetup{type=table}
\caption{\textbf{VFD scoring and threshold-calibration settings.} $J$ denotes the number of alternative observations per scored call, and $q$ denotes the quantile used to calibrate each source-specific threshold.}
\label{tab:vfd_reproduction}
\centering
\footnotesize
\setlength{\tabcolsep}{3pt}
\renewcommand{\arraystretch}{1.15}
\begin{tabular*}{\linewidth}{@{\extracolsep{\fill}}p{0.67\linewidth}l@{}}
\toprule
Setting & Value \\
\midrule
Alternative observations per scored call $J$ & 3 \\
Source-specific threshold quantile $q$ & 30\% \\
Warmup fraction of total training steps & 10\% \\
Scored eligible calls per trajectory & 1 \\
Missing-score assignment with complete call records & $z_i=1$ \\
\bottomrule
\end{tabular*}
\end{minipage}\par\addvspace{\dimexpr\intextsep\relax}

%% file: sec/table_ablation_details.tex
\providecommand{\accalls}[2]{#1\,{\color{black!65}/\,#2}}

\begin{table*}[!t]
\centering
\caption{\textbf{Main results for trajectory-cost ablations.} VFD $\rightarrow$ All-Redundant treats every eligible call as redundant, while w/o VFD Cost removes the VFD redundancy cost. Uniform Base Cost applies the base cost to all categories, while w/o Base Cost retains only the VFD redundancy cost. Bench. Avg. is the equally weighted mean accuracy across the seven benchmarks. Variants are the same as those evaluated in the main body ablations. Bold and underlined values mark the highest and second-highest accuracies in each column.}
\label{tab:ablation_details}

\begingroup
\footnotesize
\setlength{\tabcolsep}{1.3pt}
\renewcommand{\arraystretch}{1.18}

\begin{tabular*}{\textwidth}{@{\extracolsep{\fill}}l*{7}{c}c@{}}
\toprule
& \multicolumn{7}{c}{\textbf{Benchmark Acc. $\uparrow$ / Calls}}
& \textbf{Bench. Avg.} \\
\cmidrule(lr){2-8}
\cmidrule(l){9-9}
\textbf{Variant}
& HR-4K & HR-8K & V$^\star$Bench & VisualProbe & ZoomBench & MMStar
& \shortstack{MME-\\RW-Lite} & Acc.$\uparrow$ \\
\midrule
Vanilla GRPO
& \accalls{\underline{82.88}}{1.17}
& \accalls{\textbf{80.62}}{1.16}
& \accalls{\textbf{92.15}}{1.07}
& \accalls{41.36}{1.15}
& \accalls{56.64}{1.04}
& \accalls{\underline{67.47}}{1.25}
& \accalls{\underline{54.77}}{1.07}
& \underline{67.98} \\
\midrule
VFD $\rightarrow$ All-Redundant
& \accalls{81.12}{0.53}
& \accalls{78.75}{0.64}
& \accalls{89.53}{0.30}
& \accalls{\textbf{45.24}}{1.00}
& \accalls{52.61}{0.46}
& \accalls{65.13}{0.05}
& \accalls{53.36}{0.43}
& 66.53 \\
w/o VFD Cost
& \accalls{82.25}{1.26}
& \accalls{\underline{79.00}}{1.30}
& \accalls{90.05}{1.25}
& \accalls{41.17}{1.34}
& \accalls{\underline{58.18}}{1.11}
& \accalls{66.00}{1.34}
& \accalls{53.36}{1.17}
& 67.14 \\
\addlinespace[3pt]
Uniform Base Cost
& \accalls{78.25}{1.13}
& \accalls{75.50}{1.16}
& \accalls{86.91}{1.11}
& \accalls{41.36}{1.14}
& \accalls{54.98}{1.02}
& \accalls{65.93}{1.17}
& \accalls{52.53}{1.07}
& 65.07 \\
w/o Base Cost with $K=0$
& \accalls{\textbf{83.25}}{1.28}
& \accalls{75.62}{1.33}
& \accalls{\underline{91.62}}{1.16}
& \accalls{\underline{43.88}}{1.24}
& \accalls{57.23}{1.05}
& \accalls{67.07}{1.52}
& \accalls{53.10}{1.42}
& 67.40 \\
\midrule
\textbf{RA-GRPO (Ours)}
& \accalls{82.50}{1.03}
& \accalls{78.00}{1.03}
& \accalls{89.01}{1.00}
& \accalls{\underline{43.88}}{1.00}
& \accalls{\textbf{58.41}}{1.00}
& \accalls{\textbf{69.27}}{0.90}
& \accalls{\textbf{55.86}}{0.98}
& \textbf{68.13} \\
\bottomrule
\end{tabular*}
\endgroup
\end{table*}

%% file: sec/X_ablation_details.tex
\section{Ablation Study Details}
\label{sec:ablation_details}
\stepcounter{table}

Table~\ref{tab:ablation_details} presents detailed benchmark-wise ablation results. Our method achieves the highest macro-average accuracy among the evaluated variants.

\section{Additional Results on CoF-SFT}
\label{sec:cof_results}
We conduct an additional comparison using CoF-SFT-7B~\cite{chain_of_focus} to initialize RA-GRPO. CoF-SFT-7B is based on Qwen2.5-VL-7B-Instruct and uses a tool-call argument format similar to that of Qwen3-VL-8B-Instruct. We also evaluate the released CoF-SFT and CoF-RL checkpoints as reference models. Table~\ref{tab:cof_summary} summarizes benchmark-level performance and category-wise results.

\begin{table}[!htb]
\centering
\caption{Additional results on CoF. Bench. Acc. and Calls are equally weighted averages over the seven benchmarks. Beneficial and Harmful report pooled accuracy on Tool-Beneficial and Tool-Harmful samples, respectively.}
\label{tab:cof_summary}
\footnotesize
\setlength{\tabcolsep}{1.3pt}
\renewcommand{\arraystretch}{1.15}
\begin{tabular*}{\linewidth}{@{\extracolsep{\fill}}p{0.40\linewidth}rrrr@{}}
\toprule
Model & \shortstack{Bench.\\Acc.} & Calls & Beneficial & Harmful \\
\midrule
Qwen2.5-VL-7B-Instruct & 52.98 & 0.00 & 36.16 & 53.01 \\
\midrule
CoF-SFT & 59.74 & 0.52 & 50.81 & 55.42 \\
CoF-RL & 59.98 & 0.50 & 52.44 & 55.42 \\
\midrule
CoF-SFT + RA-GRPO & 62.61 & 0.00 & 58.31 & 54.22 \\
\bottomrule
\end{tabular*}
\end{table}

Training CoF-SFT with RA-GRPO on our training set improves benchmark macro-average accuracy by 2.63 percentage points over CoF-RL, but the resulting model avoids tool use entirely. Despite this behavior, it achieves 58.31\% accuracy on Tool-Beneficial samples, compared with 52.44\% for CoF-RL, while maintaining similar accuracy on Tool-Harmful samples. These results suggest that some tool calls made by CoF-RL are redundant. Because the model learns to avoid tool calls entirely, we do not further analyze the effects of RA-GRPO in this setting. This phenomenon may be associated with the low tool-call frequency of the SFT initialization and requires further investigation in future work.

%% file: sec/X_vfd_other_tools.tex
\onecolumn
\section{VFD Case Studies Beyond Cropping}
\label{sec:vfd_other_tools}
\noindent
\begin{minipage}{\textwidth}
\centering
\includegraphics[width=\textwidth,height=0.58453336\textheight,keepaspectratio]{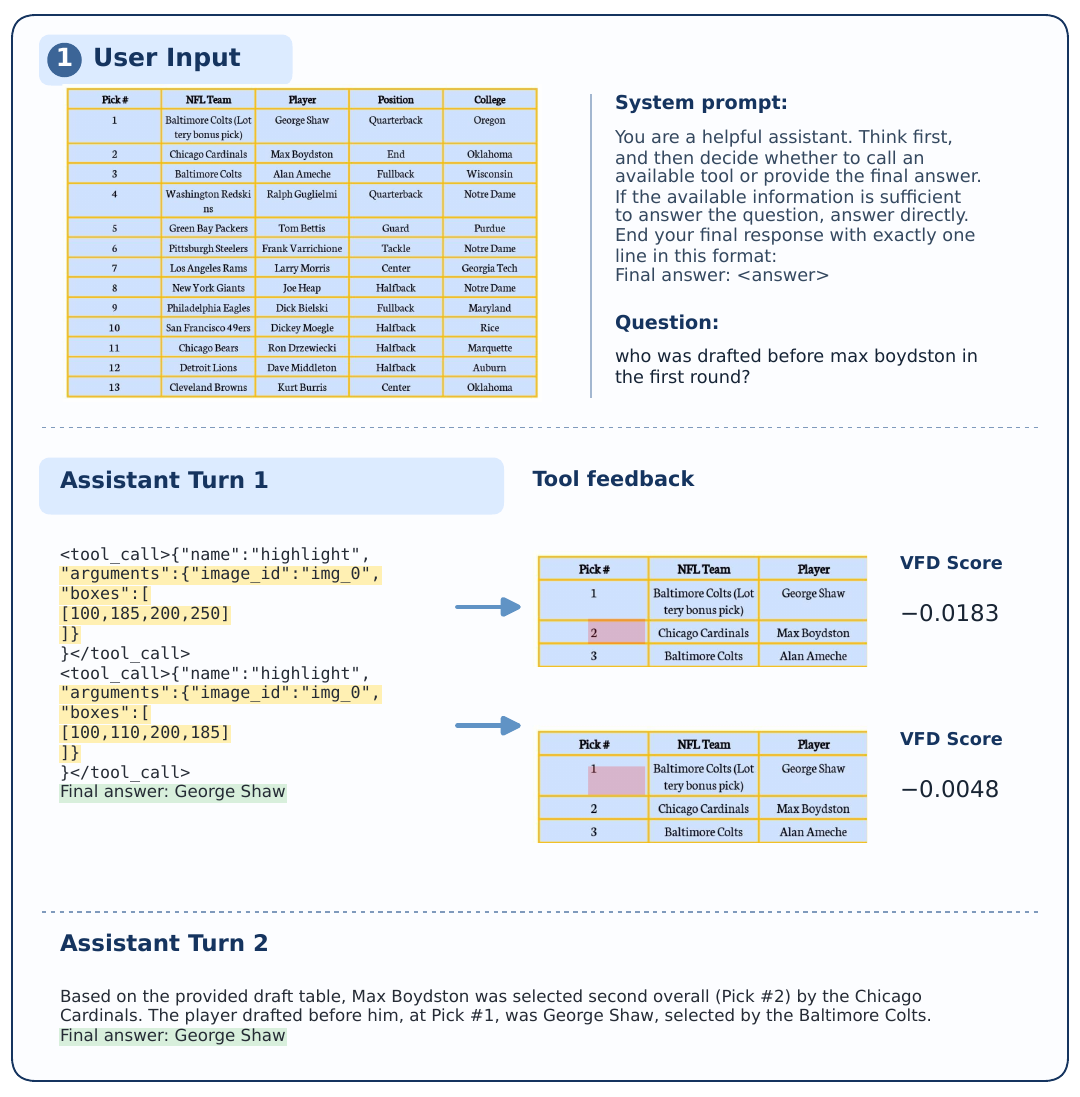}
\captionof{figure}{\textbf{Low VFD accompanies highlight calls issued alongside an already correct answer.} Assistant Turn 1 contains two highlight calls followed by the correct answer, before either tool result is returned. Assistant Turn 2 provides an explanation and repeats the answer. The two calls receive VFD scores of $-0.0183$ and $-0.0048$. Each score compares the mean log-probability of the same subsequent reasoning tokens under actual feedback and the pre-edit input image as alternative feedback, excluding the final answer and retaining all other context. These values indicate little additional support from either highlight for the subsequent reasoning. Feedback images are shown as excerpts for readability. Yellow shading marks tool arguments, and green shading marks final answers.}
\label{fig:vfd-highlight-2601}
\end{minipage}
\clearpage
\begin{center}
\includegraphics[width=\textwidth,height=0.73568314\textheight,keepaspectratio]{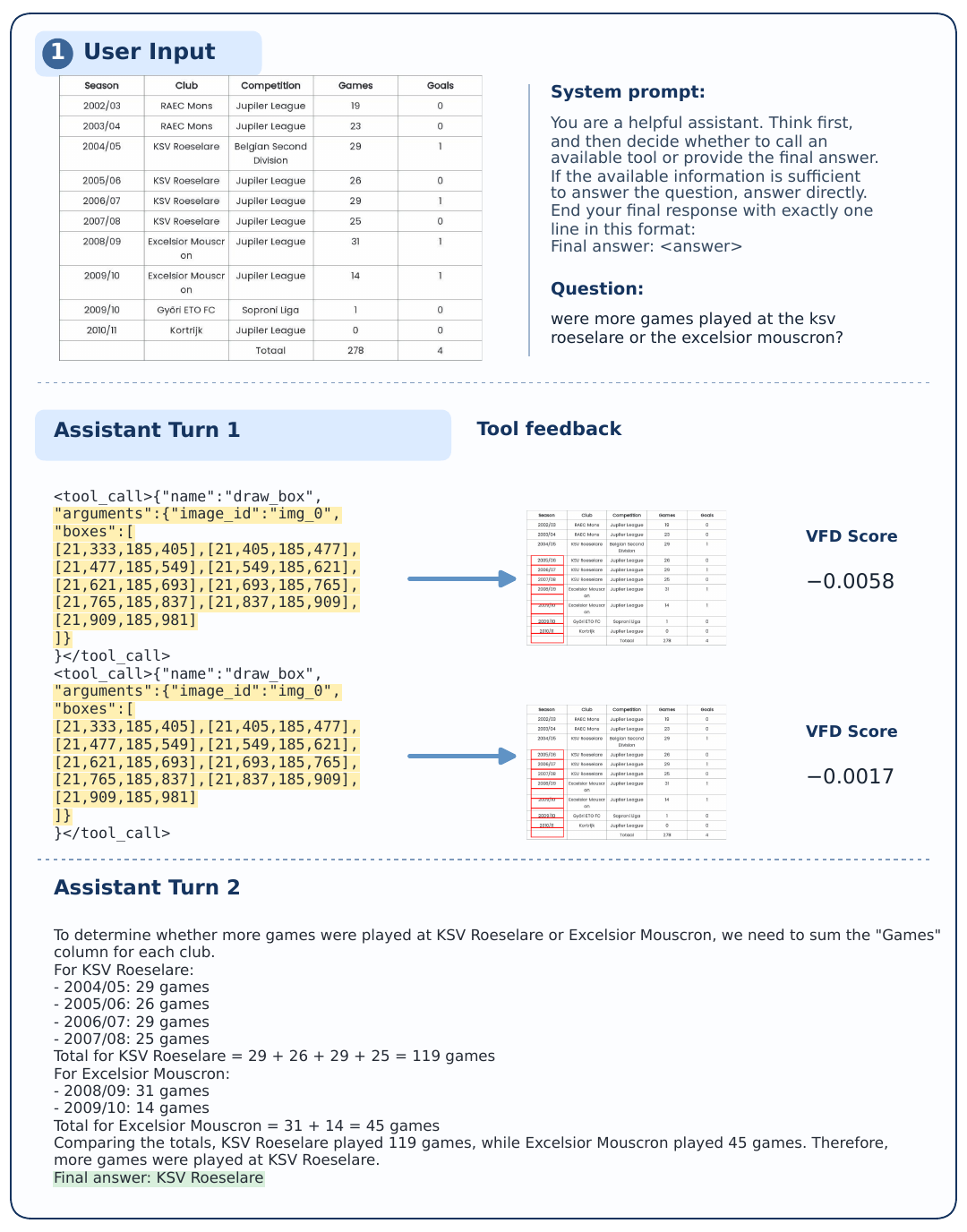}
\captionof{figure}{\textbf{Duplicate box-drawing calls yield near-zero VFD scores.} Assistant Turn 1 issues two identical \texttt{draw\_box} requests on the same input image, producing identical visual feedback. Assistant Turn 2 sums the games played for each club and gives the correct answer. The calls receive VFD scores of $-0.0058$ and $-0.0017$. Each score compares the mean log-probability of the same subsequent reasoning tokens under actual feedback and the pre-edit input image as alternative feedback, excluding the final answer. The other duplicate feedback remains available during each comparison. The scores therefore indicate limited dependence on either individual observation when its duplicate is retained. Yellow shading marks tool arguments, and green shading marks the final answer.}
\label{fig:vfd-draw-box-1503}
\end{center}
\clearpage
\begin{center}
\includegraphics[width=\textwidth,height=0.77309172\textheight,keepaspectratio]{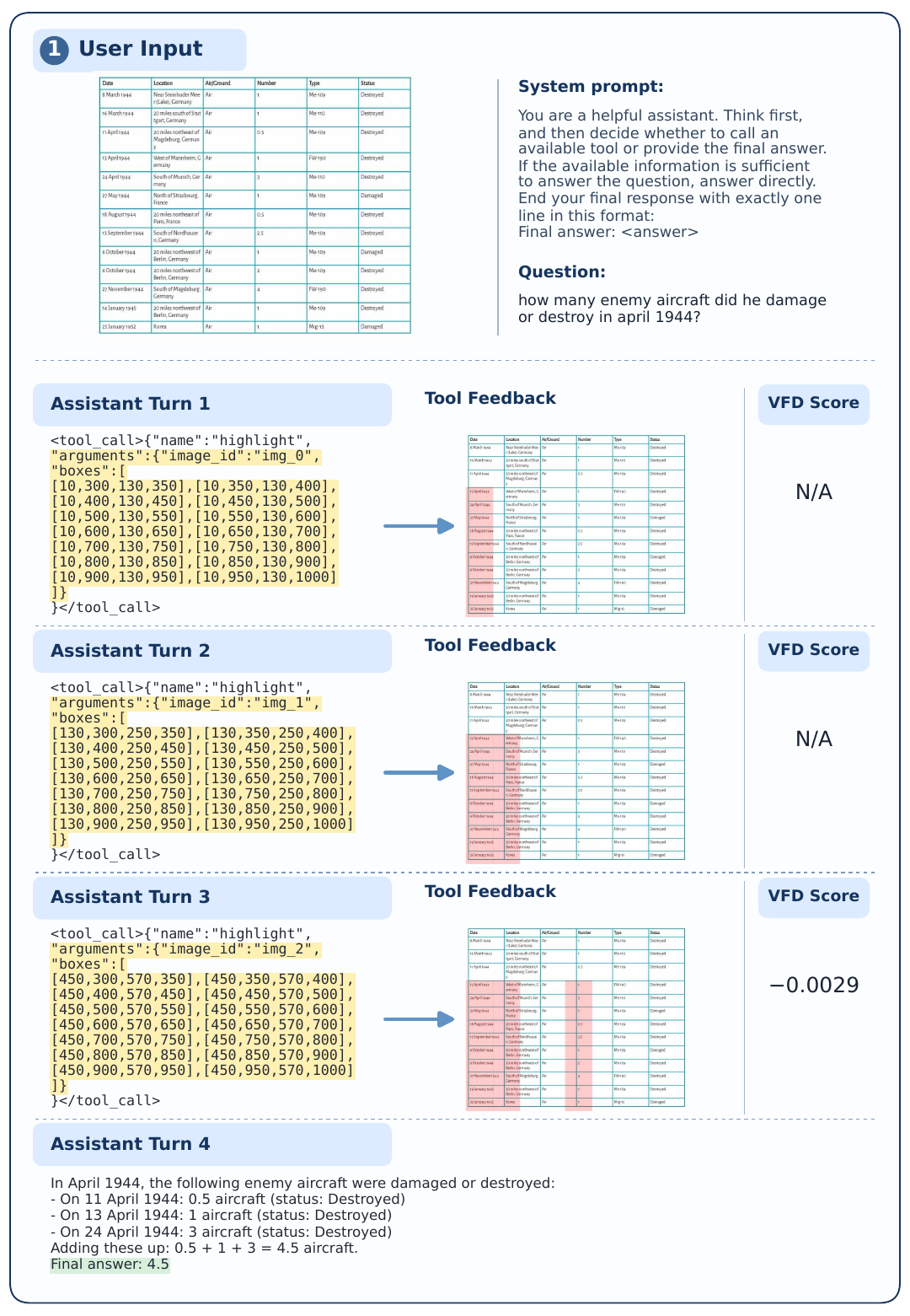}
\captionof{figure}{\textbf{VFD indicates limited additional support from the final highlight in a sequence of cumulative edits.} Assistant Turns 1--3 each issue a highlight call, starting from the original image and subsequently using the image returned by the preceding call. Assistant Turn 4 computes the total and gives the correct answer. The first two scores are marked N/A because the next assistant message contains only another tool call, leaving no intervening reasoning text to score. The third call receives a VFD score of $-0.0029$, computed over the subsequent reasoning and excluding the final answer. Replacing its feedback with the pre-edit input preserves the earlier highlights, so this score measures dependence on the final edit given the preceding visual information. Yellow shading marks tool arguments, and green shading marks the final answer.}
\label{fig:vfd-highlight-3816}
\end{center}